\documentclass{article}
\usepackage{cite}
\usepackage{graphicx} %
\usepackage{tikz}
\usetikzlibrary{decorations.pathmorphing}
\usepackage{caption}
\usepackage{subcaption}
\usepackage{amsmath}
\usepackage[capitalise]{cleveref}
\usepackage{authblk}
\usepackage{amsthm}
\usepackage{url}

\usepackage{array}
\usepackage{booktabs}

\theoremstyle{definition}
\newtheorem{exmp}{Example}[section]

\theoremstyle{definition}
\newtheorem{definition}{Definition}

\title{Directly Follows Graphs Go Predictive Process Monitoring With Graph Neural Networks}

\author[1]{Attila Lischka\thanks{ORCID: 0009-0003-1371-5996}}
\author[1]{Simon Rauch\thanks{ORCID: 0009-0008-4669-0656}}
\author[1]{Oliver Stritzel\thanks{ORCID: 0000-0003-1772-7428}}

\affil[1]{Center for Applied Research on Supply Chain Services, Fraunhofer IIS, Nuremberg, Germany}
\date{}

\begin{document}

\maketitle
\begin{abstract}
    In the past years, predictive process monitoring (PPM) techniques based on artificial neural networks have evolved as a method to monitor the future behavior of business processes. 
    Existing approaches mostly focus on interpreting the processes as sequences, so-called traces, and feeding them to neural architectures designed to operate on sequential data such as recurrent neural networks (RNNs) or transformers.
    In this study, we investigate an alternative way to perform PPM: by transforming each process in its directly-follows-graph (DFG) representation we are able to apply graph neural networks (GNNs) for the prediction tasks.
    By this, we aim to develop models that are more suitable for complex processes that are long and contain an abundance of loops.
    In particular, we present different ways to create DFG representations depending on the particular GNN we use. The tested GNNs range from classical node-based to novel edge-based architectures. Further, we investigate the possibility of using multi-graphs. By these steps, we aim to design graph representations that minimize the information loss when transforming traces into graphs.
\end{abstract}
\section{Introduction}
Predictive process monitoring (PPM) is a discipline strongly related to process mining. 
The goal of PPM is to predict the future behavior of (business) processes, such as next-activity-prediction, next-timestamp-prediction, remaining-time-prediction, process-suffix-prediction, or outcome-prediction.
By this, resources can be allocated more efficiently, deadlines can be designed robustly and errors can be prevented or detected early.
In recent years, deep-learning-based approaches using artificial neural networks have emerged as a way to tackle PPM. 
The neural networks transform the input representation of a process into a representation in a mathematical space. The network is then trained to map this mathematical representation to the desired output, such as the next activity happening in the process.

So far, existing studies dealing with deep-learning-based PPM approaches have heavily relied on representing processes as sequences, called \textit{traces} in process mining terminology. 
Due to this sequence-based representation, neural network architectures suitable for learning on sequential data have been the option of choice.
Such architectures include recurrent neural networks (RNNs) like long-short-term-memory (LSTM) (\cite{hochreiter1997long}) and transformers (\cite{vaswani2017attention}).

In this work, we take an alternative approach. 
By representing processes as graphs instead of sequences, we are able to apply neural architectures from the realm of graph learning, so-called graph neural networks (GNNs). 
We create graph representations from the processes by using the well-known concept of \textit{directly-follows-graphs (DFGs)} from the domain of process mining.
By representing processes as DFGs instead of traces, we expect better generalization performance on complex processes (e.g., with many loops) and, potentially, memory requirement benefits.
Compared to other works that have already used GNNs in PPM settings, we present a broader study that tests different GNN types (among them a novel edge-based architecture) and tackles different PPM tasks. We also investigate multi-graph-based DFG representations that reduce information loss when transforming the processes into graph representations.

The remainder of this paper is organized as follows:
We present related work in \cref{sec:relatedwork}. Then we introduce the most important background in \cref{sec:prelim} before moving on to the methodology in \cref{sec:methodology}. 
We present a small experimental evaluation in \cref{sec:experiments}. 
Finally we present our conclusion and potential future work in \cref{sec:conclusion}.

\subsection{Related Work} \label{sec:relatedwork}
An early work using LSTM networks for PPM tasks is \cite{tax2017predictive}. Similarly, \cite{camargo2019learning} is another study leveraging LSTMs to predict the continuation of processes. 
In \cite{hinkka2018exploiting}, an RNN is trained for PPM tasks using clustering to capture additional event log information as model inputs.
\cite{nolle2018binet} train a GRU \cite{cho2014learning} to detect anomalies in processes.
Instead of RNNs, \cite{bukhsh2021processtransformer} and \cite{wuyts2024sutran} use transformer-based models in their PPM settings.
Another, more recent work is \cite{roider2024efficient} where a more efficient training of an LSTM model for PPM is explored. 

One of the few papers using GNNs to perform PPM tasks is \cite{weinzierl2021exploring} where different graph representations of processes are investigated.
Another one is \cite{duong2023remaining} where the time aspects of processes are predicted using GNNs.
\cite{amiri2024pgtnet} create graphs capturing the processes by using graph nodes representing combinations of activities and life-cycle attributes before using GNNs to predict the processes' remaining time.

\subsection{Preliminaries} \label{sec:prelim}
\subsubsection{Predictive Process Monitoring}
In order to describe the most important tasks typically tackled in predictive process monitoring (PPM) settings, we first need to introduce some concepts from the domain of process mining.

\begin{definition}
    An \textit{event} $e_i = (c, a, t, r_1, \dots , r_m)$ describes a single step in an overall process. It consists of an id $c$ which indicates the specific process execution it is associated with (called \textit{trace}, see below), an activity $a$, a timestamp $t$ and, optionally, additional features $r_1, \dots, r_m$ such as personal resources. We use the notation $\#_c (e_i)$,  $\#_a (e_i)$, $\#_t (e_i)$ to refer to the id, activity and timestamp of an event.
\end{definition}
\begin{definition}
A \textit{trace} $\sigma = \langle e_1, e_2, \dots,  e_n \rangle$ is a sequence of events that describes an individual execution of a process (e.g., the production of an individual car instance). This means that all events in a trace have the same id. Further it holds that $\#_t (e_i) \leq \#_t (e_j)$ for $i < j$. 
\end{definition}
We provide an example of a trace with id $997$ in Example \ref{example:trace}. Note that, apart from the mandatory id, activity and timestamps (in hours), the events also include an additional resource feature (anonymized staff members).

\begin{exmp}
 An example trace: $\langle (997, a, \text{0:00}, r_1), (997, b, \text{0:30}, r_2),$ \newline $ (997, c, \text{0:45}, r_2), (997, b,\text{1:30}, r_1),(997, c, \text{2:45}, r_1), (997, d, \text{7:00}, r_1) \rangle$ \label{example:trace}
\end{exmp}

\begin{definition}
    An \textit{event log} $\mathcal{E}$ refers to a set of different traces $\mathcal{E} = \{ \sigma_1, \dots, \sigma_N \}$ that all describe the same process. E.g., the production process of all cars that were produced in a given period. When training a model for PPM tasks, we typically use an event log as our dataset.
\end{definition}
The particular tasks we are interested in solving using PPM in this work are the following:
\begin{itemize}
    \item \textbf{Next activity prediction:} Given a prefix $pre_i(\sigma) = \langle e_1, \dots, e_i \rangle$ of an overall trace $\sigma = \langle e_1, \dots, e_i, e_{i+1}, \dots e_n \rangle$ we want to predict the activity $\#_a(e_{i+1})$ of the next event $e_{i+1}$ in the trace $\sigma$.
    \item \textbf{Remaining time prediction:} Given a prefix $pre_i(\sigma) = \langle e_1, \dots, e_i \rangle$ of an overall trace $\sigma = \langle e_1, \dots, e_i, e_{i+1}, \dots e_n \rangle$ we want to predict the time stamp of the last event $\#_t(e_{n})$ in the trace $\sigma$.
\end{itemize}
We note that in general there are many more possible tasks, e.g., predicting the next timestamps, predicting additional event features or predicting the outcome of a trace.

\subsubsection{Directly Follows Graphs}
Another form of representing processes are directly-follows-graphs (DFGs).
Given a trace $\sigma = \langle e_1, \dots, e_n \rangle$ we create a DFG representation by defining a set of nodes $V = \{\#_a(e_i) | e_i \in \sigma \}$ that represents the activities in the trace. Further, we define a set of directed graph edges $E = \{(\#_a(e_i),\#_a(e_{i+1})) | i \in \{1, \dots, |\sigma|-1\}  \}$ that represents transitions between consecutive activities in the trace.
We note that in this work we focus on ``trace-level'' DFGs, whereas in process mining ``event-level'' DFGs are of importance where a DFG is created for a whole event log (compare, e.g., the inductive miner \cite{leemans2013discovering}).

\begin{figure}[ht]

        \centering
\begin{tikzpicture}[->,>=stealth,shorten >=1pt,auto,node distance=2.5cm,
                    thick,main node/.style={circle,draw,minimum size=1.5cm,inner sep=0pt}]

  \node[main node] (A) {a};
  \node[main node] (B) [right of=A] {b};
  \node[main node] (C) [right of=B] {c};
  \node[main node] (D) [right of=C] {d};

  \path[every node/.style={font=\sffamily\small}]
    (A) edge node {} (B)
    (B) edge[bend left] node {} (C)
    (C) edge[bend left] node {} (B)
    (C) edge node {} (D);
    
\end{tikzpicture}
\caption{DFG of the trace in Example \ref{example:trace}}
        \label{fig:exampledfg}
\end{figure}
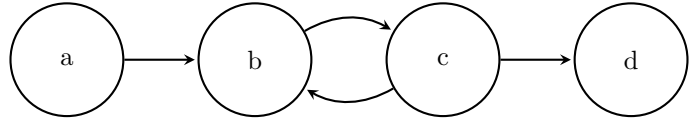

A simple DFG visualizing the trace of Example \ref{example:trace} can be found in \cref{fig:exampledfg}.

In general, DFGs offer some advantages over representing traces as sequences.
We demonstrate this by Example \ref{exmp:tires} and Example \ref{exmp:thesis}.

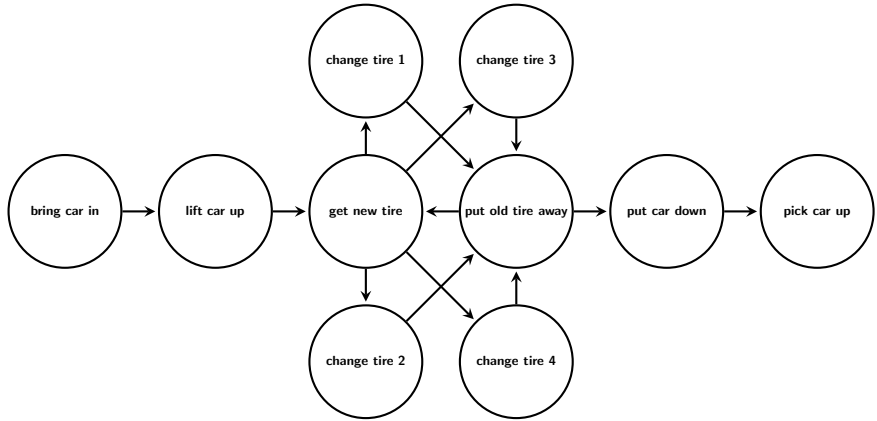
\begin{figure}[ht]

        \centering
\begin{tikzpicture}[->,>=stealth,shorten >=1pt,auto,node distance=2cm,
                    thick,main node/.style={circle,draw,minimum size=1.5cm,inner sep=0pt,font=\sffamily\tiny\bfseries}]

  \node[main node] (A) {bring car in};
  \node[main node] (B) [right of=A] {lift car up};
  \node[main node] (C) [right of=B] {get new tire};
  \node[main node] (D) [right of=C] {put old tire away};
  
  \node[main node] (E) [right of=D] {put car down};
  
  \node[main node] (F) [right of=E] {pick car up};

  \node[main node] (G) [above of=C] {change tire 1};
  \node[main node] (H) [below of=C] {change tire 2};
  \node[main node] (I) [above of=D] {change tire 3};
  \node[main node] (J) [below of=D] {change tire 4};

  \path[every node/.style={font=\sffamily\small}]
    (A) edge node {} (B)
    (B) edge node {} (C)
    (D) edge node {} (C)
    (D) edge node {} (E)
    (C) edge node {} (G)
    (C) edge node {} (H)
    (C) edge node {} (I)
    (C) edge node {} (J)

    (G) edge node {} (D)
    (H) edge node {} (D)
    (I) edge node {} (D)
    (J) edge node {} (D)
    
    (E) edge node {} (F);
\end{tikzpicture}
\caption{Process of changing tires on a car}
        \label{fig:tires}
\end{figure}

\begin{exmp}
Consider the process of changing the tires of a car.
The process consists of several steps: 
First, the car is brought into the shop and lifted up. 
Afterwards, new tires are carried to the car and the tires are switched one by one (in no particular order) and the old tires are moved out of the way. Finally, the car is put down again and picked up by the owner.
Note, that there are $4! = 24$ possible orders to change the tires, which all lead to the same result.
This means, there are 24 possible traces that capture the same process.
However, consider the DFG of this process which is shown in \cref{fig:tires}.
Note that this DFG is the exact same for all 24 trace possibilities. 
\label{exmp:tires}
\end{exmp}

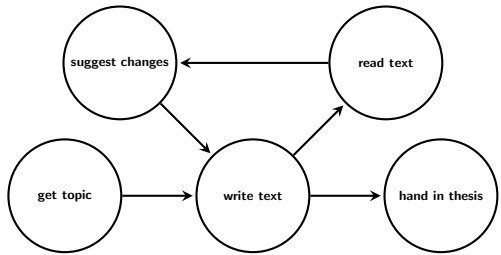
\begin{figure}[ht]

        \centering
\begin{tikzpicture}[->,>=stealth,shorten >=1pt,auto,node distance=2.5cm,
                    thick,main node/.style={circle,draw,minimum size=1.5cm,inner sep=0pt,font=\sffamily\tiny\bfseries}]

  \node[main node] (A) {get topic};
  \node[main node] (B) [right of=A] {write text};
  \node[main node] (C) [right of=B] {hand in thesis};
  \node[main node] (D) [above right of=B] {read text};
  
  \node[main node] (E) [above left of=B] {suggest changes};

  \path[every node/.style={font=\sffamily\small}]
    (A) edge node {} (B)
    (B) edge node {} (C)
    (B) edge node {} (D)
    (D) edge node {} (E)
    (E) edge node {} (B);
    
\end{tikzpicture}
\caption{Process of writing a thesis as a student, with a feedback loop for the supervisor}
        \label{fig:thesiswriting}
\end{figure}

\begin{exmp}
    Lets consider one further process: the process of a student writing a thesis. 
    In a simple model, we assume the process consist of five steps: The students receives a topic, writes some text, the supervisor reads the text and suggests changes. The student writes some new text which the supervisor reads again. This happens multiple times until the student, finally, hands the thesis in.
    Note that a trace representing this process can become arbitrarily long since, in theory, there can be arbitrary many feedback loops in the thesis writing.
    However, the DFG of the process, which we visualize in \cref{fig:thesiswriting} is of a bounded size and does not grow larger with each feedback loop.
    \label{exmp:thesis}
\end{exmp}

Based on the examples \ref{exmp:tires} and \ref{exmp:thesis} we hypothesize that DFGs can, under certain circumstances, be the better representaion of processes due to their better generalization and limit in size. 
We hypothesize this can have positive effects in the domain of PPM when using GNNs operating on DFGs instead of sequnce-based models operating on traces.

\subsubsection{Graph Neural Networks}
Graph neural networks are a group of neural architectures where, in contrast to other neural architectures like MLPs, the connections between the neurons are not arbitrary or grid-shaped but reflect the structure of the input data.
GNNs have been used successfully in many tasks dealing with graph structured data such as social networks, road networks or chemistry \cite{wu2020comprehensive}.
GNNs are related to the Weisfeiler Leman Graph Isomorphism test \cite{morris2019weisfeiler}. Like this test, GNNs iteratively compute node-features for the nodes in the input graph by aggregating information over the node neighborhoods.
Well known examples of GNNs are the graph convolutional network (GCN; \cite{kipf2016semi}) and graph attention network (GAT; \cite{velivckovic2017graph, shi2020masked}).

\begin{exmp} \label{exm:GNN}
  
Consider a graph $G = (V, E)$ where $V = \{1, \dots, n\}$ is a set of $n$ nodes and $E \subseteq V \times V $ is a set of edges between the nodes.
Then, in the $\ell$th layer of a potential GNN, the feature vector $v_{i}^\ell$ of node $i$ is computed the following way:
$$ v_{i}^{\ell} = \sigma( W^\ell_1 v_{i}^{\ell-1} + \sum_{j \in N(i)} \beta_j W^\ell_2 v_{j}^{\ell-1})$$
Here, $\sigma$ is a non-linear activation function such as ReLU, $W_1^\ell, W_2^\ell$ are learnable weight matrices of a suitable dimension and $\Sigma$ is an aggregation function such as sum. 
$N(i)$  defines the neighborhood of node $i$ and is defined as $N(i) = \{j | (j,i) \in E \}$ and $\beta_j$ is a scaling factor. In GATs, this factor is determined using the attention mechanism. In GCNs the scaling factor is based on the degree of the node $j$.
\end{exmp}

Whereas most GNNs to date are based on nodes and their associated node features, some GNNs also consider edge features when updating the node feature vectors (e.g., GAT).
In this work, we also consider a recent, purely edge-based GNN called graph edge attention network (GREAT; \cite{lischka2024greatarchitectureedgebasedgraph}).

\section{Building DFGs for GNNs} \label{sec:methodology}
In the following, we discuss how exactly the DFGs are constructed that are later on passed to the GNNs.
We design the DFGs based on the particular GNN that is used, design ``normal'' single-edge as well as multi-graph options. The different options are discussed in \cref{sec:dfgforgnns}
Further, we explain in \cref{sec:furtherdfgproc} how exactly the DFGs are built and what information the graphs encode in their node and edge features.

\subsection{Different DFG Representations} \label{sec:dfgforgnns}
To describe how the DFGs are constructed, we consider Example \ref{example:trace} again whose vanilla DFG is shown in \cref{fig:exampledfg}.

In our framework, depending on the chosen GNN, we adjust and enrich the plain DFGs with node and edge features to take advantage of the particular GNN versions.  
Moreover, we also create DFG representations that can be used with the edge-based GREAT architecture. 
Further, in addition to the ``normal'' graph representations, we also propose the usage of multi-graph DFGs.
In total, we present 6 different DFG representations for 3 different GNN versions (GCN, GAT, GREAT) in the multi-edge and single-edge cases each. We provide an overview of the different representations based on Example \ref{example:trace} in \cref{tbl:table_of_dfgs}.

We now describe the different DFGs representations: 
DFGs passed to a GCN only contain node features. The node features encode the activity the node represents, additionally, other event-features like the resources are encoded. 
However, since in our example we have activities b and c happening twice with different resources (first with resource $r_2$ and later resource $r_1$) we have to choose which resource to encode. 
In our setting, we choose the resource of the event that was last associated with the activity.
In the multi-graph representation, we have an additional edge between nodes b and c.
The DFGs used with the GAT are similar but they contain additional edge features that represent the passed time between two consecutive events. In case two activities follow each other more then once, we chose the time of the last occurrence in the single-graph case. In the multi-graph setting, we can include all occurrences.
Additionally to the DFGs for node-based GNNs like GCN and GAT, we also present an edge-based representation that can be used with GREAT.
There, we do not have any node features. 
To make up for this, we include an additional dummy start node $s$. Now, each event is represented as an edge that leads into the node of the event's activity. The edge feature further encodes information about the events resource and time. In the multi-graph case this means that we do not lose any information when encoding the trace as a graph, since even if a activity occurs repeatedly with different resources (or other associated event features) we can encode this in the corresponding edge. 
The DFGs of GREAT can be interpreted as an extension of the DFG for GAT, where some features where encoded in the edge dimensions. By taking this idea further, we transform all features into edge features and prevent information loss.

\newcommand{\dummyfigure}{\tikz \fill  (0,0) rectangle node [black] {Figure} (2,2);}
\newcommand{\gcnsingle}{
\resizebox{0.4\textwidth}{!}{ %
\begin{tikzpicture}[->,>=stealth,shorten >=1pt,auto,node distance=2.5cm,
                    thick,main node/.style={circle,draw,minimum size=1.5cm,inner sep=0pt}]

  \node[main node , label=below:{\(v_a = \begin{bmatrix} enc(a) \\ enc(r_1) \end{bmatrix}\)}] (A) {a};
  \node[main node, label=below:{\(v_b = \begin{bmatrix} enc(b) \\ enc(r_1) \end{bmatrix}\)}]  (B) [right of=A] {b};
  \node[main node, label=below:{\(v_c = \begin{bmatrix} enc(c) \\ enc(r_1) \end{bmatrix}\)}]  (C) [right of=B] {c};
  \node[main node, label=below:{\(v_d = \begin{bmatrix} enc(d) \\ enc(r_1) \end{bmatrix}\)}]  (D) [right of=C] {d};
  
  \path[every node/.style={font=\sffamily\small}]
    (A) edge node {} (B)
    (B) edge[bend left] node {} (C)
    (C) edge[bend left] node {} (B)
    (C) edge node {} (D);
    
\end{tikzpicture}
}
}

\newcommand{\gcnmulti}{
\resizebox{0.4\textwidth}{!}{ %
\begin{tikzpicture}[->,>=stealth,shorten >=1pt,auto,node distance=2.5cm,
                    thick,main node/.style={circle,draw,minimum size=1.5cm,inner sep=0pt}]
  \node[main node , label=below:{\(v_a = \begin{bmatrix} enc(a) \\ enc(r_1) \end{bmatrix}\)}] (A) {a};
  \node[main node, label=below:{\(v_b = \begin{bmatrix} enc(b) \\ enc(r_1) \end{bmatrix}\)}]  (B) [right of=A] {b};
  \node[main node, label=below:{\(v_c = \begin{bmatrix} enc(c) \\ enc(r_1) \end{bmatrix}\)}]  (C) [right of=B] {c};
  \node[main node, label=below:{\(v_d = \begin{bmatrix} enc(d) \\ enc(r_1) \end{bmatrix}\)}]  (D) [right of=C] {d};

  \path[every node/.style={font=\sffamily\small}]
    (A) edge node {} (B)
    (B) edge[bend left] node {} (C)
    (B) edge node {} (C)
    (C) edge[bend left] node {} (B)
    (C) edge node {} (D);
    
\end{tikzpicture}
}
}

\newcommand{\gatsingle}{
\resizebox{0.4\textwidth}{!}{ %
\begin{tikzpicture}[->,>=stealth,shorten >=1pt,auto,node distance=2.5cm,
                    thick,main node/.style={circle,draw,minimum size=1.5cm,inner sep=0pt}]

  \node[main node , label=below:{\(v_a = \begin{bmatrix} enc(a) \\ enc(r_1) \end{bmatrix}\)}] (A) {a};
  \node[main node, label=below:{\(v_b = \begin{bmatrix} enc(b) \\ enc(r_1) \end{bmatrix}\)}]  (B) [right of=A] {b};
  \node[main node, label=below:{\(v_c = \begin{bmatrix} enc(c) \\ enc(r_1) \end{bmatrix}\)}]  (C) [right of=B] {c};
  \node[main node, label=below:{\(v_d = \begin{bmatrix} enc(d) \\ enc(r_1) \end{bmatrix}\)}]  (D) [right of=C] {d};
  
  \path[every node/.style={font=\sffamily\small}]
    (A) edge node {0:30} (B)
    (B) edge[bend left] node {1:15} (C)
    (C) edge[bend left] node[above] {0:45} (B)
    (C) edge node {4:15} (D);
    
\end{tikzpicture}
}
}

\newcommand{\gatmulti}{
\resizebox{0.4\textwidth}{!}{ %
\begin{tikzpicture}[->,>=stealth,shorten >=1pt,auto,node distance=2.5cm,
                    thick,main node/.style={circle,draw,minimum size=1.5cm,inner sep=0pt}]

  \node[main node , label=below:{\(v_a = \begin{bmatrix} enc(a) \\ enc(r_1) \end{bmatrix}\)}] (A) {a};
  \node[main node, label=below:{\(v_b = \begin{bmatrix} enc(b) \\ enc(r_1) \end{bmatrix}\)}]  (B) [right of=A] {b};
  \node[main node, label=below:{\(v_c = \begin{bmatrix} enc(c) \\ enc(r_1) \end{bmatrix}\)}]  (C) [right of=B] {c};
  \node[main node, label=below:{\(v_d = \begin{bmatrix} enc(d) \\ enc(r_1) \end{bmatrix}\)}]  (D) [right of=C] {d};
  
  \path[every node/.style={font=\sffamily\small}]
    (A) edge node {0:30} (B)
    (B) edge[bend left] node {1:15} (C)
    (B) edge node {0:15} (C)
    (C) edge[bend left] node[above] {0:45} (B)
    (C) edge node {4:15} (D);
    
\end{tikzpicture}
}
}

\newcommand{\greatsingle}{
\resizebox{0.4\textwidth}{!}{ %
\begin{tikzpicture}[->,>=stealth,shorten >=1pt,auto,node distance=2.6cm,
                    thick,main node/.style={circle,draw,minimum size=1.4cm,inner sep=0pt}]
  \node[main node]  (S) {s};
  \node[main node]  (A) [right of=S] {a};
  \node[main node]  (B) [right of=A] {b};
  \node[main node]  (C) [right of=B] {c};
  \node[main node]  (D) [right of=C] {d};
  
  \path[every node/.style={font=\sffamily\footnotesize}]
    (S) edge node {{\(\begin{bmatrix} enc(a) \\ enc(r_1) \\ 0:00 \end{bmatrix}\)}} (A)
    (A) edge node {{\(\begin{bmatrix} enc(b) \\ enc(r_2) \\ 0:30 \end{bmatrix}\)}} (B)
    (B) edge[bend left] node {{\(\begin{bmatrix} enc(c) \\ enc(r_1) \\ 1:15 \end{bmatrix}\)}} (C)
    (C) edge[bend left] node[below] {{\(\begin{bmatrix} enc(b) \\ enc(r_1) \\ 0:45 \end{bmatrix}\)}} (B)
    (C) edge node {{\(\begin{bmatrix} enc(d) \\ enc(r_1) \\ 4:15 \end{bmatrix}\)}} (D);
    
\end{tikzpicture}
}
}

\newcommand{\greatmulti}{
\resizebox{0.4\textwidth}{!}{ %
\begin{tikzpicture}[->,>=stealth,shorten >=1pt,auto,node distance=2.6cm,
                    thick,main node/.style={circle,draw,minimum size=1.4cm,inner sep=0pt}]
  \node[main node]  (S) {s};
  \node[main node]  (A) [right of=S] {a};
  \node[main node]  (B) [right of=A] {b};
  \node[main node]  (C) [right of=B] {c};
  \node[main node]  (D) [right of=C] {d};
  \node at (8.7,-1.2){*  {\footnotesize \(\begin{bmatrix} enc(c) \\ enc(r_2) \\ 0:15 \end{bmatrix}\)}}; %
  
  \path[every node/.style={font=\sffamily\footnotesize}]
    (S) edge node {{\(\begin{bmatrix} enc(a) \\ enc(r_1) \\ 0:00 \end{bmatrix}\)}} (A)
    (A) edge node {{\(\begin{bmatrix} enc(b) \\ enc(r_2) \\ 0:30 \end{bmatrix}\)}} (B)
    (B) edge[bend left] node {{\(\begin{bmatrix} enc(c) \\ enc(r_1) \\ 1:15 \end{bmatrix}\)}} (C)
    (B) edge node {*} (C)
    (C) edge[bend left] node[below] {{\(\begin{bmatrix} enc(b) \\ enc(r_1) \\ 0:45 \end{bmatrix}\)}} (B)
    (C) edge node {{\(\begin{bmatrix} enc(d) \\ enc(r_1) \\ 4:15 \end{bmatrix}\)}} (D);
    
\end{tikzpicture}
}
}

\newcolumntype{M}[1]{>{\centering\arraybackslash}m{#1}}

\begin{table}
    \centering
    \begin{tabular}{cM{47mm}M{47mm}}
       \toprule
         & single-graph & multi-graph  \\
        \midrule
        GCN & \gcnsingle & \gcnmulti  \\
        GAT & \gatsingle & \gatmulti  \\
        GREAT & \greatsingle & \greatmulti \\
        \bottomrule
    \end{tabular}
    \caption{Overview of DFGs for different GNNs}
    \label{tbl:table_of_dfgs}
\end{table}

\subsection{Further DFG Processing} \label{sec:furtherdfgproc}
We now describe what additional adjustments and encodings are made to the DFGs. Further, we explain how they are processed by the GNNs and finally used to make predictions.

\subsubsection{Adjustments and technicalities}

When building DFGs, we decided to include nodes for all activities in a training dataset in a trace's DFG, even if an activity is not part of a particular trace (resulting in a disconnected node). Further, we add self-loop-edges for all nodes. As a preparation for potential future ``trace generation'' tasks where whole traces shall be predicted, we add additional dummy start and end tokens such that the models trained on the traces can also learn to predict the first event and when a trace is complete (similar to the ``$\left[ \text{END} \right]$'' tokens in NLP contexts).

Encodings for categorical variables like activities and resources are made by using embedders that map the categories to high dimensional embeddings.
In the case of GCN and GAT, node embeddings further encode how often each activity occurs in the trace in total. 
Similarly, we encode in the single-graph GAT and GREAT settings how often certain edges occurred. Also, regarding time features in single-edge graphs, we encode the duration of the last time this edge was taken additionally to the average and maximum time of this edge.
Furthermore, we encode with flags if a node/edge has been part of the the suffix of the last 5 events in the trace and which one corresponds to the very last event.

\subsubsection{GNNs for PPM Pipeline}
The DFGs that were built using the procedure described so far are afterwards passed to the corresponding GNN models.
As was shown in Example \ref{exm:GNN}, GNNs compute hidden feature vectors for the nodes in the graph.
After several layers (w.l.o.g. $\ell$), we have hidden feature vectors $v_i^\ell$ for all nodes $i$ in the DFG.
We can then apply pooling (e.g. summation) to get a feature representation for the whole graph $v_G = \sum_{i \in V} v_i^\ell$.
By concatenating the representations of the last activity in a trace (w.l.o.g $z$) and the whole graph, we can get a representation of the whole trace that incorporates global and local information $v_{trace} = (v_G || v_z^\ell)$.
This trace vector can then be fed into an MLP and used for the remaining time prediction and next activity prediction (or potential other PPM tasks).

\section{Experiments} \label{sec:experiments}
For each GNN we conduct experiments for the next activity prediction and the remaining time prediction task on the Helpdesk \footnote{\url{https://data.4tu.nl/articles/_/12675977}}, Hospital Billing\footnote{\url{https://data.4tu.nl/articles/_/12705113}} and BPIC2012\footnote{\url{https://data.4tu.nl/articles/_/12689204/1}} datasets. We experiment on both: multi- and single-graph DFGs.

The results of the next activity task are given in \cref{tbl:nextact} in form of accuracy and f-score metrics.
The results of the remaining time task can be found in \cref{tbl:remtime}. There, we present mean absolute error (MAE) and root mean squared error (RMSE).

The results are achieved by training 5-layer GNNs with a hidden dimension of 64. Models are implemented in PyTorch \cite{ansel2024pytorch} and PyTorch Geometric \cite{Fey/Lenssen/2019}. We use the NAdam optimizer \cite{dozat2016incorporating} for training with a learning rate of $0.0001$. Each dataset is split into 80\% training, 10\% validation and 10\% test data. 
We train for a maximum of 100 epochs and use early stopping in case there is no improvement on the validation dataset for 10 epochs.
Afterwards, the models are evaluated on the test datasets. The results presented in this section correspond to the model performance on these test datasets.

Note that for comparison we also report the performance of a re-implementation of the LSTM in \cite{camargo2019learning} for the next activity task and the transformer in \cite{bukhsh2021processtransformer} for the next activity and remaining time tasks. 
For the re-implementations, we stay as close to the original architectures as possible. As a result, the LSTM is not trained for the remaining time prediction task and the transformer in the remaining time prediction tasks is trained using days ($d$) as the unit compared to hours ($h$) that we use for our GNNs. For easier comparison, we transform the results of the trained model to hours as well. We further note that \cite{bukhsh2021processtransformer} only has one transformer layer and the \cite{camargo2019learning} has two LSTM layers. Therefore, the sequence-based models have fewer parameters than our 5-layer deep GNNs.

\begin{table}[h]
    \centering
    \begin{tabular}{cccc}
        \cline{2-4}
        & Helpdesk & Hospital & BPIC2012  \\
        \cline{2-4}
        & Acc. / F-Score & Acc. / F-Score & Acc. / F-Score \\
        \cline{2-4} 
        LSTM \cite{camargo2019learning} & 0.842 / 0.811  & 0.913 / 0.893 & 0.786 / 0.718\\
        \hline
        Transformer \cite{bukhsh2021processtransformer} & 0.843 / 0.813  & 0.819 / 0.777 & 0.714 / 0.624 \\
        \hline
        GCN single &  0.854 / 0.826 & 0.931 / 0.921 & 0.870 / 0.849\\
        GCN multi & 0.857 / 0.829 & 0.931 / 0.921 & 0.871 / 0.849 \\
        \hline
        GAT single & \underline{0.862} / 0.834 & \textbf{0.933} / 0.922 & \textbf{0.875} / 0.864 \\
        GAT multi & \textbf{0.862} / 0.835 & 0.931 / 0.922 & \underline{0.874} / 0.862\\
        \hline
        GREAT single & 0.860 / 0.832 & 0.932 / 0.921  & 0.868 / 0.849 \\
        GREAT multi & 0.852 / 0.824 & \underline{0.932} / 0.922 & 0.862 / 0.851\\

    \end{tabular}
    \caption{Next Activity Prediction}
    \label{tbl:nextact}
\end{table}

\begin{table}[h]
    \centering
    \begin{tabular}{ccccc}
        \cline{3-5}
        & & Helpdesk & Hospital & BPIC2012  \\
        \cline{2-5}
        & unit & MAE / RMSE & MAE / RMSE & MAE / RMSE \\
        \cline{2-5}
        Transformer \cite{bukhsh2021processtransformer} & \textit{d} & 5.46 / 7.32 & 57.90 / 104.33 & 6.67 / 11.22   \\
        Transformer \cite{bukhsh2021processtransformer} & \textit{h} & \textbf{131.11} / 175.61 & 1389.54 / 2503.96 & 160.17 / 269.33   \\
        \hline
        GCN single & \textit{h} & 168.16 / 206.91 & 1780.57 / 2762.72 & 149.02 / 250.53   \\
        GCN multi & \textit{h} & 168.82 / 209.38 & 1784.10 / 2766.28 &   149.22 / 250.59 \\
        \hline
        GAT single & \textit{h} & 151.27 / 193.10 &  \underline{1281.35} / 2368.18 &  \textbf{131.91} / 237.27   \\
        GAT multi & \textit{h} & \underline{150.84} / 211.88 & \textbf{1241.41} / 2318.30 &   \underline{132.28} / 237.99 \\
        \hline
        GREAT single & \textit{h} & 245.77 / 316.75 & 1314.87 / 2277.34& 168.12 / 264.27  \\
        GREAT multi & \textit{h} & 463.71 / 499.06 & 1412.90 / 2401.58 & 180.02 / 262.93 \\

    \end{tabular}
    \caption{Remaining Time Prediction}
    \label{tbl:remtime}
\end{table}

\subsection{Results}

On the next activity prediction task, GAT generally achieves the best performance on all datasets. However, we note that the other GNNs perform almost equally well, outperforming the sequence-based models.

On the remaining time prediction task, the performance between the different models fluctuates more noticeably. 
While the transformer-based model performs best on the Helpdesk dataset (followed by the GAT), GAT performs best on the other datasets (followed by single-edge GREAT on the Hospital dataset and GCN on BPIC2012).

Generally using multi-graphs compared to single-graphs does not seem to influence performance significantly. Only for GREAT it results in noticeable differences in the remaining time prediction tasks.
We hypothesize that this limited impact on performance is due to our traces not being overly complicated and not having many loops and duplicate edges influencing the outcomes.

\section{Conclusion and Future Work} \label{sec:conclusion}
In this study, we investigate the idea of using GNNs for PPM tasks using DFG representations of process traces.
We compare several ways to enrich and adjust DFGs for three different GNN types by including different node- and edge-based features.
We are the first ones to make the problem completely edge-based and consider multi-graph settings, which allows us to restrict the information loss when transforming traces into DFG representations.
In an experimental evaluation, we benchmark the different approaches against each other.

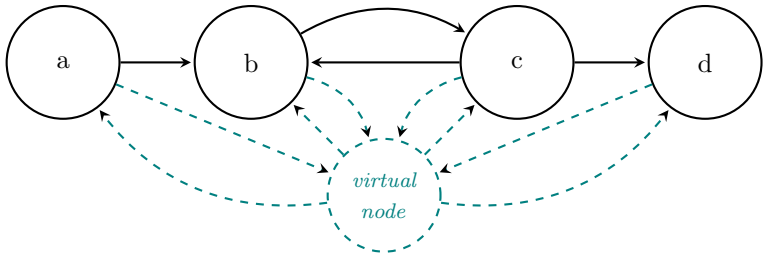
\begin{figure}[ht]

        \centering
\begin{tikzpicture}[->,>=stealth,shorten >=1pt,auto,node distance=2.5cm,
                    thick,main node/.style={circle,draw,minimum size=1.5cm,inner sep=0pt}]

  \node[main node] (A) {a};
  \node[main node] (B) [right of=A] {b};
  \node[main node,draw=teal, dashed, text width=1cm,align=center, text=teal] (V) [below right of=B] {\footnotesize \textit{virtual node}};
  \node[main node] (C) [above right of=V] {c};
  \node[main node] (D) [right of=C] {d};

  \path[every node/.style={font=\sffamily\small}]
    (A) edge node {} (B)
    (B) edge[bend left] node {} (C)
    (C) edge node {} (B)
    (C) edge node {} (D)
    (V) edge[draw=teal, bend left, dashed] node {} (A)
    (V) edge[draw=teal, dashed] node {} (B)
    (V) edge[draw=teal, dashed] node {} (C)
    (V) edge[draw=teal, bend right, dashed] node {} (D)
    (A) edge[draw=teal, dashed] node {} (V)
    (B) edge[draw=teal, bend left, dashed] node {} (V)
    (C) edge[draw=teal, bend right, dashed] node {} (V)
    (D) edge[draw=teal, dashed] node {} (V);
    
\end{tikzpicture}
\caption{DFG of the trace in Example \ref{example:trace} with additional virtual node}
        \label{fig:virtualnode}
\end{figure}

In an extension of this work, we aim to tackle existing limitations of the proposed approach.
Most importantly, we want to introduce virtual nodes in the DFG representations \cite{hwang2022analysis}.
These allow a fast information flow between all graph components. Compare \cref{fig:virtualnode} (an adaption of Example \ref{example:trace} and \cref{fig:exampledfg}) where all nodes can immediately share information by passing it through the virtual node, wheres information can only travel one node at a time in \cref{fig:exampledfg}.

Further, we want to do an in-depth analysis of the advantages of GNNs operating on DFGs compared to RNNs (or other sequence-based models) on trace representations. 
For this, we want to find event logs that provide characteristics favorable of the strengths of both approaches. 
We hypothesize that DFGs will be advantageous on long traces with repeating activities.
Such traces can get arbitrarily long, whereas the DFGs would generally remain of compact size.
Potentially we will investigate generating suitable synthetic event logs with existing tools for event log generation (compare \cite{maldonado2024gedi}).

\bibliographystyle{plain} %
\bibliography{literature}

\begin{thebibliography}{10}

\bibitem{amiri2024pgtnet}
Keyvan Amiri~Elyasi, Han van~der Aa, and Heiner Stuckenschmidt.
\newblock Pgtnet: A process graph transformer network for remaining time prediction of business process instances.
\newblock In {\em International Conference on Advanced Information Systems Engineering}, pages 124--140. Springer, 2024.

\bibitem{ansel2024pytorch}
Jason Ansel, Edward Yang, Horace He, Natalia Gimelshein, Animesh Jain, Michael Voznesensky, Bin Bao, Peter Bell, David Berard, Evgeni Burovski, et~al.
\newblock Pytorch 2: Faster machine learning through dynamic python bytecode transformation and graph compilation.
\newblock In {\em Proceedings of the 29th ACM International Conference on Architectural Support for Programming Languages and Operating Systems, Volume 2}, pages 929--947, 2024.

\bibitem{bukhsh2021processtransformer}
Zaharah~A Bukhsh, Aaqib Saeed, and Remco~M Dijkman.
\newblock Processtransformer: Predictive business process monitoring with transformer network.
\newblock {\em arXiv preprint arXiv:2104.00721}, 2021.

\bibitem{camargo2019learning}
Manuel Camargo, Marlon Dumas, and Oscar Gonz{\'a}lez-Rojas.
\newblock Learning accurate lstm models of business processes.
\newblock In {\em Business Process Management: 17th International Conference, BPM 2019, Vienna, Austria, September 1--6, 2019, Proceedings 17}, pages 286--302. Springer, 2019.

\bibitem{cho2014learning}
Kyunghyun Cho, Bart Van~Merri{\"e}nboer, Caglar Gulcehre, Dzmitry Bahdanau, Fethi Bougares, Holger Schwenk, and Yoshua Bengio.
\newblock Learning phrase representations using rnn encoder-decoder for statistical machine translation.
\newblock {\em arXiv preprint arXiv:1406.1078}, 2014.

\bibitem{dozat2016incorporating}
Timothy Dozat.
\newblock Incorporating nesterov momentum into adam.
\newblock 2016.

\bibitem{duong2023remaining}
Le~Toan Duong, Louise Trav{\'e}-Massuy{\`e}s, Audine Subias, and Christophe Merle.
\newblock Remaining cycle time prediction with graph neural networks for predictive process monitoring.
\newblock In {\em Proceedings of the 2023 8th International Conference on Machine Learning Technologies}, pages 95--101, 2023.

\bibitem{Fey/Lenssen/2019}
Matthias Fey and Jan~E. Lenssen.
\newblock Fast graph representation learning with {PyTorch Geometric}.
\newblock In {\em ICLR Workshop on Representation Learning on Graphs and Manifolds}, 2019.

\bibitem{hinkka2018exploiting}
Markku Hinkka, Teemu Lehto, and Keijo Heljanko.
\newblock Exploiting event log event attributes in rnn based prediction.
\newblock In {\em International Symposium on Data-Driven Process Discovery and Analysis}, pages 67--85. Springer, 2018.

\bibitem{hochreiter1997long}
S~Hochreiter.
\newblock Long short-term memory.
\newblock {\em Neural Computation MIT-Press}, 1997.

\bibitem{hwang2022analysis}
EunJeong Hwang, Veronika Thost, Shib~Sankar Dasgupta, and Tengfei Ma.
\newblock An analysis of virtual nodes in graph neural networks for link prediction.
\newblock In {\em The First Learning on Graphs Conference}, 2022.

\bibitem{kipf2016semi}
Thomas~N Kipf and Max Welling.
\newblock Semi-supervised classification with graph convolutional networks.
\newblock {\em arXiv preprint arXiv:1609.02907}, 2016.

\bibitem{leemans2013discovering}
Sander~JJ Leemans, Dirk Fahland, and Wil~MP Van Der~Aalst.
\newblock Discovering block-structured process models from event logs-a constructive approach.
\newblock In {\em Application and Theory of Petri Nets and Concurrency: 34th International Conference, PETRI NETS 2013, Milan, Italy, June 24-28, 2013. Proceedings 34}, pages 311--329. Springer, 2013.

\bibitem{lischka2024greatarchitectureedgebasedgraph}
Attila Lischka, Jiaming Wu, Morteza~Haghir Chehreghani, and Balázs Kulcsár.
\newblock A great architecture for edge-based graph problems like tsp, 2024.

\bibitem{maldonado2024gedi}
Andrea Maldonado, Christian~MM Frey, Gabriel~Marques Tavares, Nikolina Rehwald, and Thomas Seidl.
\newblock Gedi: Generating event data with intentional features for benchmarking process mining.
\newblock In {\em International Conference on Business Process Management}, pages 221--237. Springer, 2024.

\bibitem{morris2019weisfeiler}
Christopher Morris, Martin Ritzert, Matthias Fey, William~L Hamilton, Jan~Eric Lenssen, Gaurav Rattan, and Martin Grohe.
\newblock Weisfeiler and leman go neural: Higher-order graph neural networks.
\newblock In {\em Proceedings of the AAAI conference on artificial intelligence}, volume~33, pages 4602--4609, 2019.

\bibitem{nolle2018binet}
Timo Nolle, Alexander Seeliger, and Max M{\"u}hlh{\"a}user.
\newblock Binet: multivariate business process anomaly detection using deep learning.
\newblock In {\em International Conference on Business Process Management}, pages 271--287. Springer, 2018.

\bibitem{roider2024efficient}
Johannes Roider, Dario Zanca, and Bjoern~M Eskofier.
\newblock Efficient training of recurrent neural networks for remaining time prediction in predictive process monitoring.
\newblock In {\em International Conference on Business Process Management}, pages 238--255. Springer, 2024.

\bibitem{shi2020masked}
Yunsheng Shi, Zhengjie Huang, Shikun Feng, Hui Zhong, Wenjin Wang, and Yu~Sun.
\newblock Masked label prediction: Unified message passing model for semi-supervised classification.
\newblock {\em arXiv preprint arXiv:2009.03509}, 2020.

\bibitem{tax2017predictive}
Niek Tax, Ilya Verenich, Marcello La~Rosa, and Marlon Dumas.
\newblock Predictive business process monitoring with lstm neural networks.
\newblock In {\em Advanced Information Systems Engineering: 29th International Conference, CAiSE 2017, Essen, Germany, June 12-16, 2017, Proceedings 29}, pages 477--492. Springer, 2017.

\bibitem{vaswani2017attention}
A~Vaswani.
\newblock Attention is all you need.
\newblock {\em Advances in Neural Information Processing Systems}, 2017.

\bibitem{velivckovic2017graph}
Petar Veli{\v{c}}kovi{\'c}, Guillem Cucurull, Arantxa Casanova, Adriana Romero, Pietro Lio, and Yoshua Bengio.
\newblock Graph attention networks.
\newblock {\em arXiv preprint arXiv:1710.10903}, 2017.

\bibitem{weinzierl2021exploring}
Sven Weinzierl.
\newblock Exploring gated graph sequence neural networks for predicting next process activities.
\newblock In {\em International conference on business process management}, pages 30--42. Springer, 2021.

\bibitem{wu2020comprehensive}
Zonghan Wu, Shirui Pan, Fengwen Chen, Guodong Long, Chengqi Zhang, and S~Yu Philip.
\newblock A comprehensive survey on graph neural networks.
\newblock {\em IEEE transactions on neural networks and learning systems}, 32(1):4--24, 2020.

\bibitem{wuyts2024sutran}
Brecht Wuyts, Seppe~Vanden Broucke, and Jochen De~Weerdt.
\newblock Sutran: an encoder-decoder transformer for full-context-aware suffix prediction of business processes.
\newblock In {\em 2024 6th International Conference on Process Mining (ICPM)}, pages 17--24. IEEE, 2024.

\end{thebibliography}
\end{document}